\pdfoutput=1

\documentclass[11pt]{article}
 
\usepackage[final]{ACL2023}

\usepackage{times}
\usepackage{latexsym}
\usepackage{multirow}
\usepackage[T1]{fontenc}
\usepackage[utf8]{inputenc}
\usepackage{graphicx}
\usepackage{microtype}
\usepackage{amsfonts}
\usepackage{inconsolata}
\usepackage{amsmath}
\usepackage{stmaryrd}
\usepackage{natbib}
\usepackage{listings}
\usepackage{booktabs}
\usepackage{makecell}
\usepackage{enumitem}
\title{Improve Language Model and Brain Alignment via Associative Memory}

\author{Congchi Yin \quad Yongpeng Zhang \quad Xuyun Wen \quad Piji Li$^{\ast}$\\ 
\textsuperscript{\rm 1} College of Artificial Intelligence, Nanjing University of Aeronautics and Astronautics\\
\textsuperscript{\rm 2} The Key Laboratory of Brain-Machine Intelligence Technology, Ministry of Education\\
\texttt{\{congchiyin,pjli\}@nuaa.edu.cn}}

\DeclareUnicodeCharacter{0301}{\'{e}}

\begin{document}
\maketitle
\renewcommand{\thefootnote}{\fnsymbol{footnote}}
\footnotetext[1]{Corresponding author.}
\renewcommand{\thefootnote}{\arabic{footnote}}
\begin{abstract}
Associative memory engages in the integration of relevant information for comprehension in the human cognition system.
In this work, we seek to improve alignment between language models and human brain while processing speech information by integrating associative memory.
After verifying the alignment between language model and brain by mapping language model activations to brain activity, the original text stimuli expanded with simulated associative memory are regarded as input to computational language models.
We find the alignment between language model and brain is improved in brain regions closely related to associative memory processing. 
We also demonstrate large language models after specific supervised fine-tuning better align with brain response, by building the \textit{Association}\footnote{\url{https://github.com/lemonsis/Association}} dataset containing 1000 samples of stories, with instructions encouraging associative memory as input and associated content as output. 
\end{abstract}

\section{Introduction}

Human language comprehension is a complicated process widely involving multiple brain functions \citep{10GESCHWIND, ABOITIZ1997381}. Previous studies \citep{dronkers2007paul,binder2015wernicke} have confirmed that Wernicke's area and Broca's area are essential in speech comprehension and language production. More relevant regions are found and subdivided to match corresponding functions through functional Magnetic Resonance Imaging (fMRI) scans in later work \citep{poremba2004species,GOUREVITCH20081,10.1162/jocn.2010.21466}. Among all the functions related to language comprehension, \textit{associative memory} \citep{anderson2014human} plays an indispensable role, serving as the key to linking together related concepts and pieces of information.

\begin{figure}[!t]
\centering
\includegraphics[width=\linewidth]{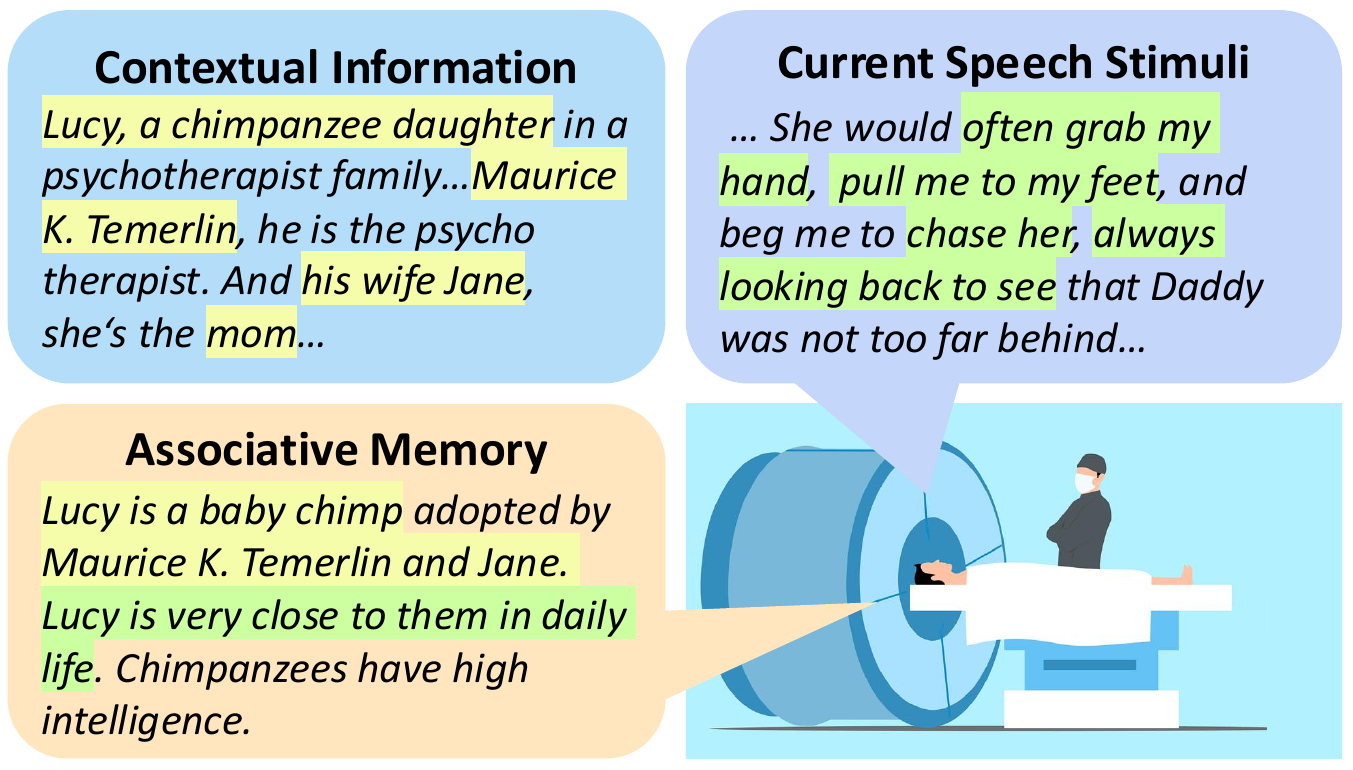}
\caption{An example of how associative memory works when subject listens to speech.}
\label{f1}
\end{figure}


The human associative memory system \citep{mayes2007associative,eichenbaum2017time} is responsible for encoding, monitoring, and retrieving diverse components of information, including basic perceptions (i.e. semantic memory \citep{binder2011neurobiology}), personal experiences (i.e. episodic memory \citep{tulving1972episodic}), and contextual details (i.e. working memory \citep{baddeley2003working}). 
While associative memory integrates multiple outside stimuli (visual, auditory, sensory, etc.), we primarily focus on associative memory during human language comprehension in this study, particularly in a task involving passively listening to continuous speech.
The core effect of associative memory in language comprehension is to help build connections between diverse concepts. Human is capable of retrieving relative concepts that facilitate language understanding instinctively \citep{schank1972conceptual}.
Although associative memory involves complex formation and interaction within the biological brain, at a high level it can be considered as the integration of associated concepts \citep{mcnamara2009toward}.
For example, as shown in Figure \ref{f1}, the orange box is what associative memory involves. The content in the purple box indicates the speech stimuli that the subject is receiving. The blue box represents previous phonetic stimuli that the subject heard a few moments ago. 

Large language models like GPT-4 \citep{achiam2023gpt} have shown remarkable natural language understanding and generation ability. They follow a next word prediction pattern, which is similar to human's manner of processing text information \citep{caucheteux2023evidence,antonello2024predictive}. Previous research \citep{jain2018incorporating,toneva2019interpreting,caucheteux2020language,goldstein2022shared} has confirmed the activations of language models can be linearly mapped to the activity of human brain when receiving the same text stimuli. 
This finding provides a powerful tool for investigating the alignment between biological brain and language models. For example, \citet{caucheteux2023evidence} showed such alignment can be improved by introducing future words prediction. \citet{moussa2024improving} fine-tuned speech model with brain-relevant semantics to improve its alignment to brain activity. However, as far as we know, few studies explore the associative memory in human language processing.

In this paper, we investigate two research questions: (1) Will simulating associative memory in brain language processing improve the alignment between language models and the human brain? (2) Can we improve the alignment between language model and brain by instructing language models to generate associative content?
We design the following experiment steps to answer these questions. First, the alignment between language model activation and human brain activity (i.e. brain score) is evaluated when they receive the same text stimuli as input. Following previous studies \citep{caucheteux2020language}, traditional language model GPT-2 \citep{radford2019language} is selected. We also try large language model LLaMA-2 \citep{touvron2023llama} for comparison. 
For the first research question, associative memory is considered as the integration of context and associated knowledge, as the example shown in Figure \ref{f1}. 
Data augmentation with simulated associative memory is performed to the original text stimuli. The activation of language model with augmented sentences as input is mapped to the original brain activity.
The brain score of regions related to associative memory (e.g. medial temporal lobe (MTL)) is recorded in comparison to the original brain score.
For the second research question, we build an instruction tuning dataset \textit{Association} containing 1000 samples with story paragraphs and instructions encouraging associative memory as input, associated content as output. \textit{Association} is applied in the supervised fine-tuning (SFT) of base large language model. The brain score of language model after SFT is re-evaluated to see whether the score of relevant regions is improved. Such improvement will become strong evidence suggesting the alignment between large language models and human brain can be improved by instructing language models to generate associative content.

Our contributions can be summarized as follows:
\begin{itemize}[topsep=0pt]
\setlength\itemsep{-0.5em}
    \item We find associative memory simulation via data augmentation is capable of improving language model and brain alignment.
    \item We release \textit{Association}, an instruction tuning dataset containing 1000 samples for investigating associative memory by supervised fine-tuning large language models.
    \item We demonstrate that fine-tuning a large language model with instructions that promote associative memory can enhance its alignment with brain activity.
\end{itemize}

\section{Related work}
\paragraph{Language Models and Brain Alignment} Previous studies have mapped word-level embeddings to fMRI or MEG signals \citep{mitchell2008predicting, huth2012continuous, huth2016natural}. \citet{jain2018incorporating,toneva2019interpreting,goldstein2022shared,caucheteux2022deep} indicated that human brain combines information of previous words to predict next words and such prediction is increasingly contextual along the hierarchy by extracting activations from different layers in language models. Such prediction has also been proven to span multiple timescales\citep{goldstein2020thinking, caucheteux2023evidence}. \citet{antonello2024scaling} further analyzed the mapping of large language models to the human brain.
Some studies seek to find the reasons behind the alignment between language models and human brain.
\citet{caucheteux2021disentangling} factorized language model activations into lexical, compositional, syntactic and semantic representations.
\citet{wehbe2014simultaneously,oota2024joint} investigated the specific linguistic properties and brain regions that contribute to such alignment. 

\begin{figure*}[!t]
\centering
\includegraphics[width=\linewidth]{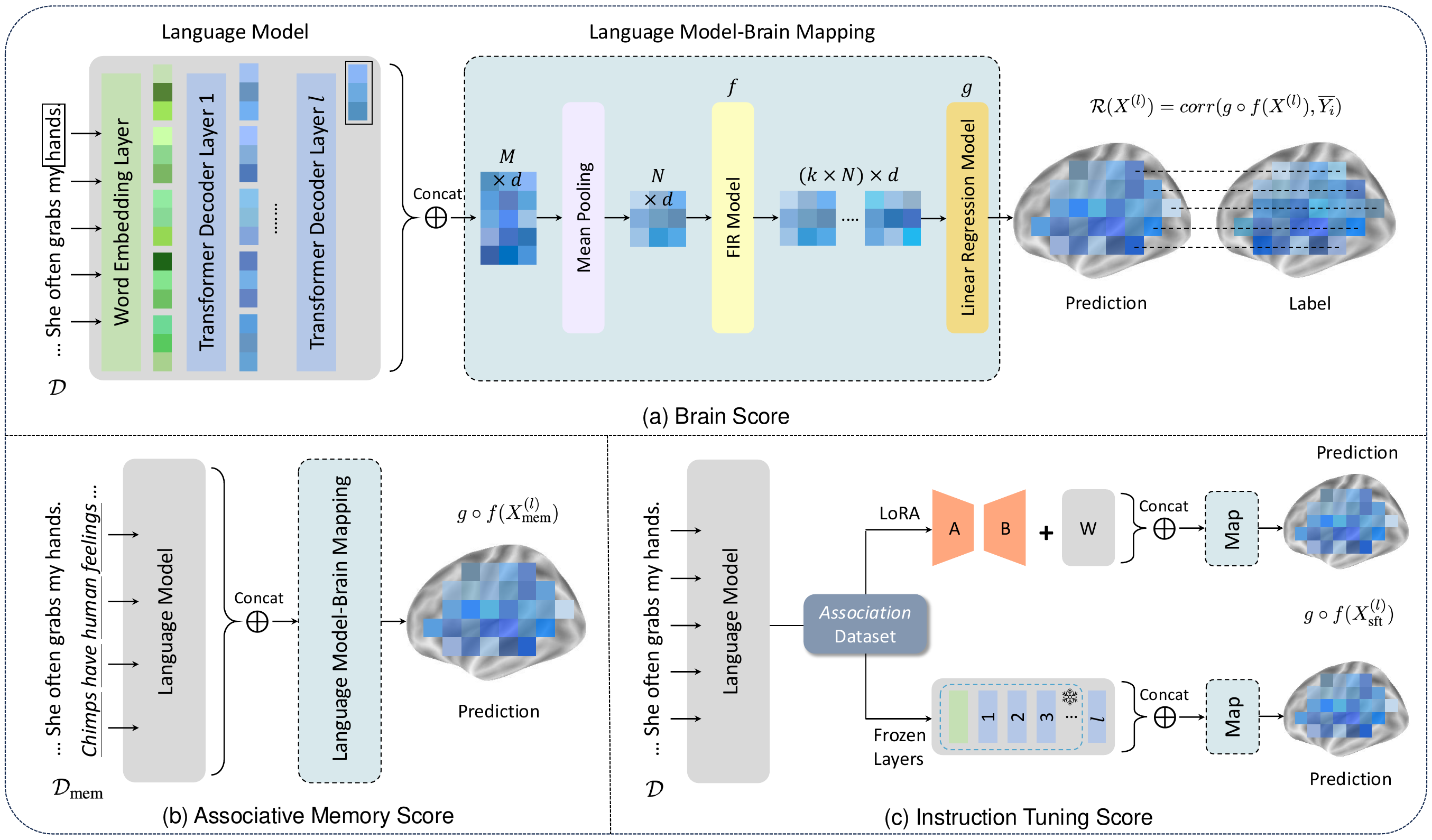}
\caption{General framework of calculating brain score, associative memory score, and instruction tuning score.}
\label{f2}
\end{figure*}

\paragraph{Associative Memory} It's a fundamental cognitive process enabling the linking of related information \citep{anderson2014human}. Early research \citep{marr1991simple} laid the groundwork by proposing theoretical models that describe how the hippocampus could facilitate the storage and retrieval of associative memories. Subsequent empirical studies \citep{mcclelland1995there} demonstrated the importance of synaptic plasticity and the role of long-term potentiation (LTP) in associative learning and memory consolidation. With neuroimaging techniques, some studies have identified key brain regions involved in associative memory, including the medial temporal lobe, the prefrontal cortex, and their interactions \cite{sperling2001encoding}. Moreover, computational models \citep{bogacz2003comparison} have been developed to simulate associative memory, providing a deeper understanding of how neurons could support complex associative tasks.

\section{Methods}
\subsection{Overview}
We first introduce the method of evaluating language model and brain alignment by mapping language model activations to fMRI signals. The extent of alignment is referred as \textbf{brain score}. Then we expand the original dataset with simulated associative memory, and recalculate the brain score to identify newly activated brain regions. Finally, an instruction tuning dataset \textit{Association} is proposed and applied in the supervised fine-tuning (SFT) of large language model (LLM). LLM after SFT is re-evaluated on the original dataset to explore whether brain score is improved by instructing large language models to generate associative content.

\subsection{Brain Score Calculation}
We aim to investigate the alignment between language models and human brain when they process text information. To better map language models to brain, auto-regressive models with left-to-right attention are selected, as brain can't get access to future information like bidirectional Transformers. The activations of language models are mapped to the fMRI recordings with the same text stimuli as input. More precisely, given a sequence $S=(s_1,\dots,s_{i},\dots,s_M)$ of $M$ words from dataset $\mathcal{D}$, the output embedding $x_i$ of $s_i$ in the $l$-th layer of language model can be written as
\begin{equation}
    \label{1}
    x_i^{(l)}=W_{l}W_{l-1}\cdots W_0(s_{i-c},\ldots,s_{i-1},s_i),
\end{equation}
where $W_{l}$ indicates the transformation weight matrix of $l$-th Transformer layer, $c$ is a hyper-parameter deciding the length of history information fused into current word embedding. The representation of word sequence $S$ through language model is denoted as $X^{(l)}=concat(x_1,\ldots,x_M)\in \mathbb{R}^{M\times d}$.

Let $Y=\Psi(S)\in \mathbb{R}^{N\times v}$ be the brain activity elicited by the same word sequence $S$, which is collected through $N$ continuous fMRI frames. $v$ is the number of voxels in brain. Analysis is conducted for one particular voxel $Y_i\in \mathbb{R}^{N}$ because it can be easily extended to whole brain. Since fMRI signals are inherently noisy, the average blood-oxygen-level-dependent (BOLD) signal $\overline{Y_i}$ across total $T$ subjects for each voxel is considered.
\begin{equation}
    \overline{Y_i}= \frac{1}{T} \sum_{j=1}^{T} Y_{ij} 
\end{equation}

As fMRI is sampled discretely with fixed time intervals (a.k.a. TR) and the sampling frequency is usually much lower than word rate, we take the mean pooling of language model activations to match $N$ fMRI frames, as shown in Figure \ref{f2}. To mitigate the gap of delayed BOLD responses, we follow previous work \citep{huth2016natural,affolter2020brain2word} and apply a finite impulse response (FIR) model. For fMRI frame $i\in \left [ 1\ldots N \right ]$, the temporal transformation $f_i$ is formally defined as
\begin{equation}
\begin{aligned}
f_i: \mathbb{R}^{N \times d} & \rightarrow \mathbb{R}^{k\times d} \\
x & \mapsto concat(\widetilde{x_i}, \widetilde{x}_{i-1}, \ldots, \widetilde{x}_{i-k+1}) 
\end{aligned}
\end{equation}
where 
\begin{equation}
\begin{aligned}
\widetilde{x_i} & =\frac{1}{m}\sum_{\substack{m \in \llbracket 1 \ldots M \rrbracket \\
\mathcal{T}(m)=i}} x_m^{(l)},
\end{aligned}
\end{equation}
\begin{equation}
\begin{aligned}
\mathcal{T}: \llbracket 1 \ldots M \rrbracket & \rightarrow \llbracket 1 \ldots N \rrbracket \\
m & \mapsto \min _{k \in \llbracket 1 \ldots N \rrbracket}\left|t_{y_k}-t_{x_m}\right|,
\end{aligned}
\end{equation}
with $\widetilde{x}$ taking the mean pooling of word embeddings between successive fMRI TRs, $k$ a hyper-parameter controlling the delay of FIR feature $x$, $(t_{x_1},\ldots,t_{x_M})$ the timings of words onsets and $(t_{y_1}, \ldots, t_{y_N})$ the timings of $N$ fMRI frames.

After achieving temporal alignment between $X^{(l)}$ and $\overline{Y_i}$ through $f$, we seek to find a linear model $g \in \mathbb{R}^d$ to map language model activations to brain activity. Ridge regression with $\ell_2$-regularization is learned to predict brain activity:
\begin{equation}
\underset{g}{\operatorname{argmin}} \sum_{i \in I_{\text {train }}}\left(\overline{Y_i}-g^T f(X^{(l)})\right)^2+\lambda\|g\|^2.
\end{equation}
Finally, similar to previous work \citep{yamins2016using}, brain score $\mathcal{R}(X^{(l)})$ is defined as correlation between predicted brain activity and original brain activity. Pearson correlation score $corr(\cdot,\cdot)$ is applied to measure such connection and brain score of each voxel can be written as
\begin{equation}
\mathcal{R}(X^{(l)}) = corr(g \circ f(X^{(l)}), \overline{Y_i}).
\end{equation}

Moreover, we design a novel brain score ceiling test to explore the limitation of explainable and predictable brain signals. In each iteration, all the subjects hearing the same word sequence are randomly separated into two parts, part $A$ and part $B$. Instead of using language model activations to predict brain activity, one part of subjects' brain activity $Y_A$ are used to predict the other part of subjects' brain activity $Y_B$ through linear model $g$. All the brain activity is averaged across corresponding subjects to reduce noise. The brain score ceiling for $i$-th voxel of brain is calculated as
\begin{equation}
    \mathcal{R}_{\text{ceiling}} = corr(g(\overline{Y_{A_i}}),\overline{Y_{B_i}}).
\end{equation}

\begin{figure*}[!t]
\centering
\includegraphics[width=\linewidth]{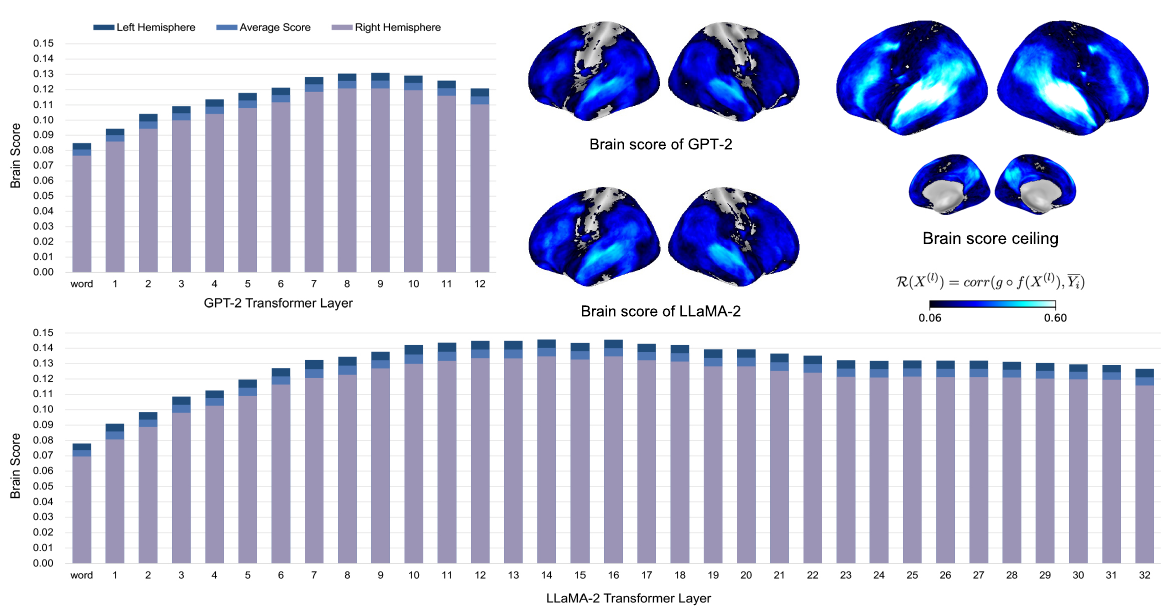}
\caption{Brain score of different layers with visualization. The colorbar refers to value of brain score.}
\label{f3}
\end{figure*}

\subsection{Data Augmentation with Simulated Associative Memory}
The concrete mechanism of associative memory in the biological brain is complex, involving the interaction of neurons from multiple brain regions.
To investigate whether the alignment between language model and human brain can be improved by associative memory, we don't directly simulate its process in the brain.
Instead, we concretize the content of associative memory, namely what people may associate during a passively story hearing test, as natural language input to language models.

The original dataset $\mathcal{D}$ is expanded with simulated associative memory. The dataset after data augmentation is denoted as $\mathcal{D}_{\text{mem}}$. Specifically, sentence-level and word-level associative memory simulation is tried respectively. Sentence-level data augmentation involves grammatically complete sentences, typically focusing on a single aspect of association. In contrast, word-level data augmentation includes words or phrases that capture multiple aspects of associative memory. Both human and GPT-4 annotations are applied. Human annotators are asked to write down what they associate when receiving certain text stimuli that trigger associative memory.
GPT-4 is not able to decide when and where to add associative memory content like human, so we give clear instructions and let GPT-4 return associated words or sentences based on the context and its knowledge every four sentences of the original text stimuli.
Examples of different data augmentation methods are shown in Appendix \ref{sec:appendix}. Considering the latency of fMRI signals, all the expanded content is put at the end of the sentences that trigger associative memory. Onsets are all set to the same as the last word's offset, as if the associative memory forms simultaneously when subject receives specific text stimuli.
 
Word sequence $S_{\text{mem}}\in \mathcal{D}_{\text{mem}}$ is used to compute the activation of language model $X_{\text{mem}}^{(l)}$. Brain activity maintains $\overline{Y_i}$ as the subjects listen to original dataset $D$.
Following the same process as mentioned before, brain score with simulated associative memory is computed through
\begin{equation}
\mathcal{R}(X_{\text{mem}}^{(l)}) = corr(g \circ f(X_{\text{mem}}^{(l)}), \overline{Y_i}).
\end{equation}
The associative memory score $\mathcal{F}$ of one specific voxel is defined as the difference between brain score with associative memory and original brain score
\begin{equation}
\mathcal{F}(X^{(l)}) = \mathcal{R}(X_{\text{mem}}^{(l)}) - \mathcal{R}(X^{(l)}).
\end{equation}

\subsection{Instruct LLM to Generate Associative Content}
Different from language models with limited parameters like GPT-2, recent large language models with huge number of parameters can be trained to follow instructions through supervised fine-tuning. We build an instruction tuning dataset \textit{Association} containing paragraphs of stories with instructions encouraging associative memory as input, word-level associated content as answers. More details about the dataset are introduced in Appendix \ref{dataset} and examples are shown in Appendix \ref{sec:appendix}. We build the \textit{Association} dataset to investigate whether the alignment between language models and human brain can be improved by instructing large language model to generate associative content. The improvement is reflected by observing the increment of brain score on certain brain regions.

Two supervised fine-tuning methods are tried: low-rank adaptation (LoRA) \citep{hu2021lora} and frozen layers finetuning. LoRA applies two trainable low-rank matrices $B\in \mathbb{R}^{d\times r}$ and $A \in \mathbb{R}^{r\times k}$ during fine-tuning and the original weight $W\in \mathbb{R}^{d\times k}$ of LLM is frozen. When the training finishes, the original weight is replaced as $W+BA$. For frozen layers fine-tuning, layers before the $l$-th layer are frozen during supervised finetuning. All the parameters in layer $l$ and layers after $l$-th layer are trainable. The weight matrix of $l$-th Transformer layer after supervised finetuning is denoted as $W_{\text{sft}}^{(l)}$ and its output is denoted as $X_{\text{sft}}^{(l)}$. Following previous methods, brain score of supervised fine-tuned model is $\mathcal{R}(X^{(l)}_{\text{sft}})$. We define instruction tuning score $\mathcal{M}$ as the growth percentage of supervised finetuned model compared to base model:
\begin{equation}
\mathcal{M}(X^{(l)}) = (\mathcal{R}(X^{(l)}_{\text{sft}})-\mathcal{R}(X^{(l)})) / \mathcal{R}(X^{(l)}).
\end{equation}

\begin{figure*}[!t]
\centering
\includegraphics[width=\linewidth]{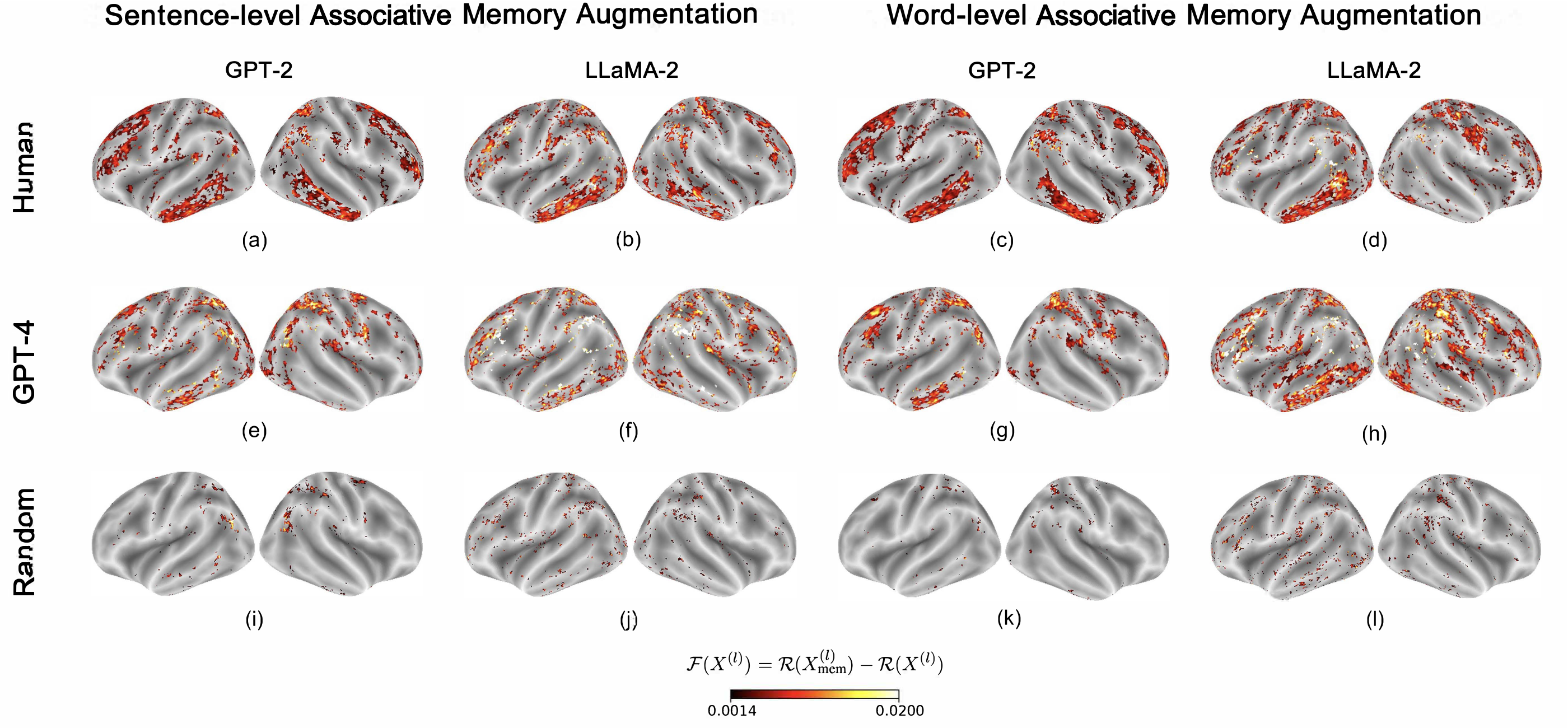}
\caption{Associative memory score of sentence-level and word-level data augmentation.}
\label{f4}
\vspace{-3mm}
\end{figure*}

\section{Experimental Setups} \label{setup}
\subsection{Datasets}
We use the publicly accessible ``Narratives'' dataset \citep{nastase2021narratives} which contains fMRI recordings of 345 individuals listening to 27 spoken English stories. After filtering short articles, 15 stories with corresponding fMRI images are selected for experiments. More details are in Appendix \ref{dataset}.

\subsection{Cortical Parcellation}

Nine brain regions are selected to analyze brain score changes in different regions of interests (ROIs). Besides inferior temporal gyrus, inferior temporal sulcus and middle temporal gyrus that are known to contribute to associative memory, we also explore regions related to speech processing and working memory, as associative memory involves interaction with working memory in a storying hearing task. These regions include middle and superior frontal gyrus, inferior and superior frontal sulcus, superior parietal lobule, angular gyrus. Details are presented in Appendix \ref{cort}.

\section{Results and Analysis}

\begin{figure*}[!t]
\centering
\includegraphics[width=\linewidth]{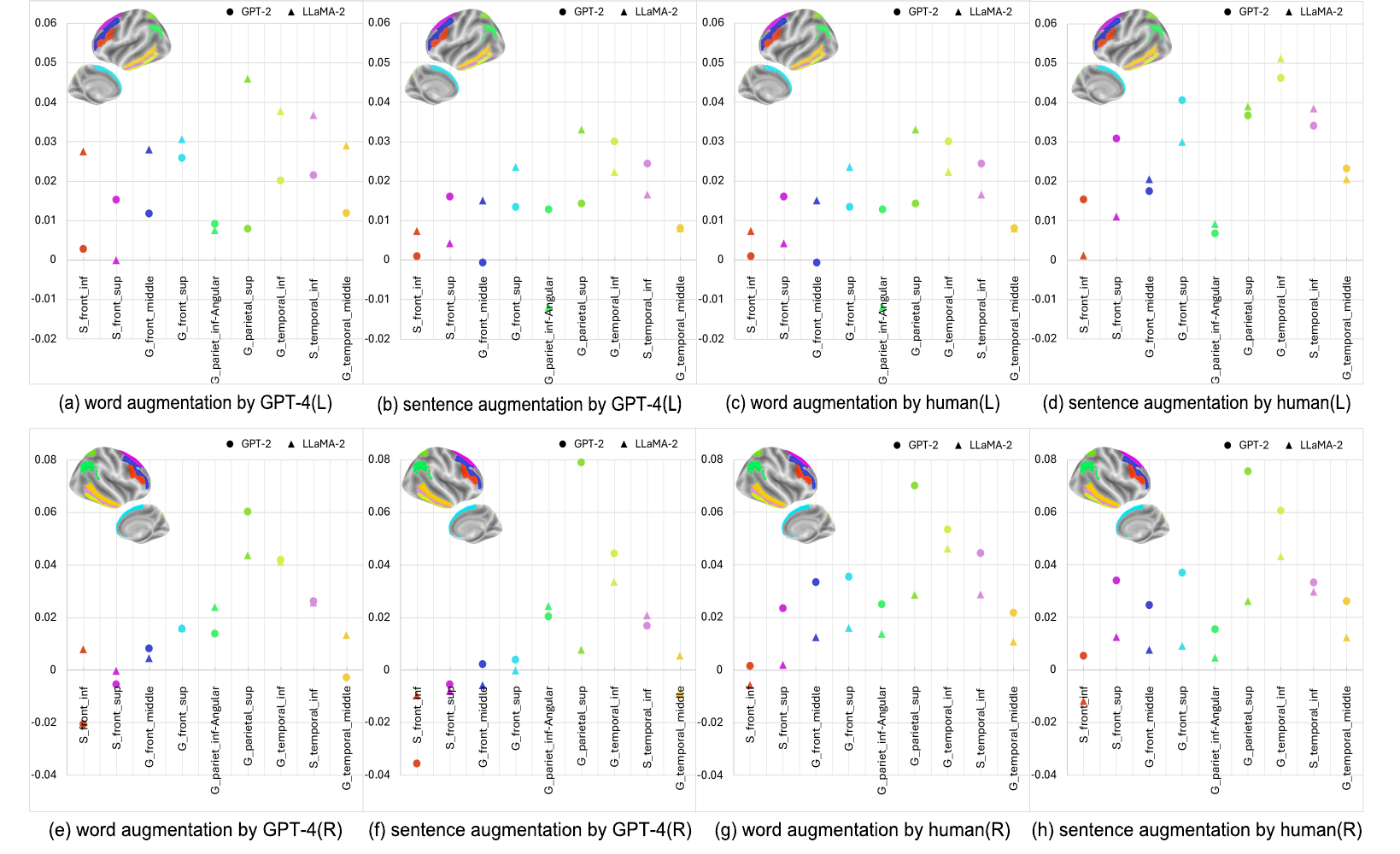}
\caption{Associative memory score of specific regions of interests (ROIs). (L) and (R) refer to left hemisphere and right hemisphere, respectively. The color of dot corresponds to the color of specific ROI.}
\label{f5}
\end{figure*}

First, we analyze the brain score of different layers for GPT-2 and LLaMA-2 models. Second, dataset with associative memory augmentation is applied to explore activated brain regions. Finally, LLaMA-2 fine-tuned in an instruction dataset \textit{Association} is evaluated to verify the improved alignment between language models and human brain.

\subsection{Brain Score Comparison}
The brain score is calculated by averaging all the fMRI voxel of all the test subjects. Two kinds of language model embeddings are considered: non-contextual word embedding and contextual embedding of each Transformer layer. Results are shown in Figure \ref{f3}. Overall, LLaMA-2 gets a higher brain score than GPT-2 due to larger number of parameters, more training corpus, and better representation ability. Brain score of left hemisphere is higher than that of right hemisphere for different layers of both models. It's consistent with previous researches \citep{halpern2005lateralization,ries2016choosing} on lateralization of brain function that Broca's and Wernicke's areas related to the production and comprehension of speech are found exclusively on the left hemisphere. Scores calculated through word embedding are significantly lower than other layers for lack of contextual information. Brain scores of both hemispheres peak at the ninth layer for GPT-2 model, achieving an average score of $0.126$, which satisfies previous conclusion \citep{caucheteux2022brains} that the activations of $l=n_{\text{layers}}\times 2/3$ layer best fit brain activity. However, we find layer best predicting brain activation becoming shallow for LLaMA-2. The fourteenth out of thirty-two layers reaches the highest brain score of $0.146$ and $0.135$ for left and right hemispheres, respectively. Relative studies \citep{durrani-etal-2021-transfer,sajjad-etal-2022-analyzing,zhang2023emergent} on investigating representation inside Transformer-based language models reveal that lower layers are dominated by lexical concepts, whereas middle and higher layers better represent core-linguistic concepts. But why middle front layers best aligned with the brain still remains unexplored. Based on the above findings, we apply the ninth layer and fourteenth layer of GPT-2 and LLaMA-2 separately for all the following experiments.

We map each voxel's brain score of GPT-2 and LLaMA-2 to brain surface and plot figures for better visualization. Figure \ref{f3} also shows the brain maps of the highest possible brain score (i.e. brain score ceiling) under current dataset and linear regression model. Brain scores are witnessed over a distributed and bilateral cortical network, peaking in middle and superior temporal gyrus, middle and superior temporal sulcus, as well as in the supra-marginal and the infero-frontal cortex.

\subsection{Associative Memory Score}

Associative memory score measures the difference between original brain score and brain score with simulated associative memory.
We investigate the associative memory score under various settings. Results are shown in Figure \ref{f4}. 
Sub-figures (a) to (d) show associative memory score with human annotated associative memory augmentation, while sub-figures (e) to (h) are with GPT-4 augmented associative memory. Besides, we apply random word-level and sentence-level data augmentation as a control group to demonstrate that the improvement of brain score benefits from associative memory. Results are shown in Sub-figures (i) to (l).
The random augmentation is conducted in the following manner. For word-level augmentation, we apply GPT-4 to generate 100 unrepeated verbs, nouns, adjectives, and randomly select words among the set of a total of 300 words. For sentence-level augmentation, we directly apply GPT-4 to randomly generate sentences.
Sub-figures (a), (b), (e), (f), (i), (j) show sentence-level associative memory augmentation, and sub-figures (c), (d), (g), (h), (k), (l) show word-level augmentation. 

Generally speaking, large and continuous regions of the brain, including some areas of frontal gyrus, frontal sulcus and parietal lobule gain increase in brain score ranging from $0.0014$ to $0.02$. Since these regions get a relatively low brain score without simulated associative memory stimulation, such a gain in brain score is considerable.
Moreover, we find random data augmentation leads to none and even negative growth of brain score, which supports the improvement of alignment is caused by introducing associative memory.
From Figure \ref{f4}, it's noticed that word-level augmentation leads to better performance compared to sentence-level augmentation on both models. We think compared to sentence-level augmentation, word-level augmentation probably benefits from multi-aspect association with less introduced noise like proposition and conjunction. Nouns, adjectives, and verbs contain more intensive information.
This finding is also consistent with previous neuroscience study \citep{schwering2020verbal}, which indicates that associative memory is conceptualized by the unit of the word.
Human annotation earns higher associative memory score than GPT-4, because GPT-4 can't decide where to generate associative content like human annotators. LLaMA-2 performs better than GPT-2 model under most cases with wider activated brain regions and higher score. Overall, the alignment between two tested language models and human brain gets significantly improved with word-level human-annotated associative memory. 

\begin{figure}[!t]
\centering
\includegraphics[width=\linewidth]{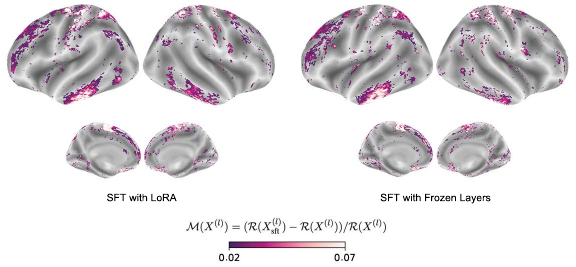}
\caption{Instruction tuning score after supervised finetuning with two different methods.}
\label{f6}
\vspace{-3mm}
\end{figure}

We also compute associative memory score on brain regions of interests (ROIs) and the results are shown in Figure \ref{f5}. Sub-figure (a) to (d) shows the cases of left hemisphere and sub-figure (e) to (h) shows right hemisphere. Since areas related to associative memory in language comprehension mainly distribute in the left hemisphere \citep{smith1998components}, results on figure (a) to (d) are more confident. 
Improvements in brain scores were observed across nine regions of interest (ROIs) associated with associative memory, as well as in areas related to speech processing and working memory, ranging from $0$ to $0.05$. Such trend of improvement is generally consistent with each of the nine ROIs for both models. LLaMA-2 model gets a higher associative memory score than GPT-2 model for most ROIs in the left hemisphere. Superior and middle frontal gyrus, superior parietal lobule related to working memory, inferior and superior frontal sulcus related to speech processing, medial temporal lobe (MTL) area related to associative memory all get improved on both word-level and sentence-level augmentation dataset.

\subsection{Instruction Tuning Score}

\begin{figure}
\centering
\includegraphics[width=\linewidth]{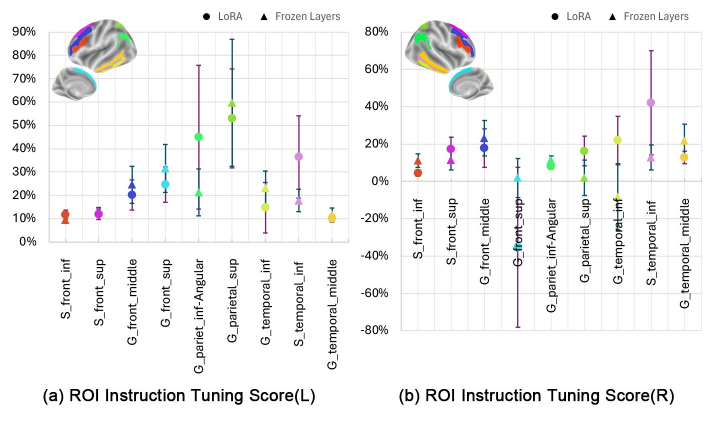}
\caption{Subject-level instruction tuning score of specific regions of interests (ROIs).}
\label{f7}
\vspace{-3mm}
\end{figure}

Common instruction tuning will not lead to improvement of brain score \citep{gao2023rolesscalinginstructiontuning}. We explore whether brain score can be improved by instructing language model to simulate associative memory. The \textit{Association} dataset is built, which contains 1000 training samples with story paragraphs and prompts encouraging associative memory as input, associated content as output. We try two different supervised fine-tuning methods, LoRA and frozen layers finetuning, for LLaMA-2 and the results are shown in Figure \ref{f6} and \ref{f7}.

As shown in Figure \ref{f6}, LLaMA-2 after supervised fine-tuning with both methods shows $2\%$ to $7\%$ gain in regions related to associative memory (i.e. medial temporal lobe (MTL)), which indicates the alignment between language model and brain is improved by associative memory instructed tuning. We also calculate instruction tuning score in specific ROIs and results are shown in Figure \ref{f7}. Different from Figure \ref{f6} where instruction tuning score $\mathcal{M}$ is computed by averaging all the subjects' voxel, which is equivalent to viewing all the subjects' brain activity as one, in Figure \ref{f7} we first calculate instruction tuning score for each subject and then average these scores. Results correspond to $95\%$ confidence intervals (CIs) across all test subjects are shown in the figure. Superior parietal lobule related to working memory gets the highest score of $50\%$ to $60\%$. As to associative memory, instruction tuning score in medial temporal lobe (i.e. G\_temporal\_inf, S\_temporal\_inf, G\_temporal\_middle) also get significantly improved across hundreds of subjects.

\section{Conclusion}
In this paper, we explore whether the alignment between language model and human brain could be improved by introducing associative memory in a passive story hearing task. By defining brain score, associative memory score, and instruction tuning score in experiments, we answer two research questions: The alignment between language model and human brain can be improved (1) with simulated associative memory (2) by instructing language models to generate associative content.

\section*{Limitations}
The ``Narratives'' dataset contains fMRI recordings when subjects listen to English spoken stories. Language and cultural background of the participants and the story should be considered. Therefore, the results could not fully cover all types of languages and cultures. Moreover, annotators involved in associative memory data augmentation may possess different language and cultural backgrounds compared to the subjects in ``Narratives'' dataset. Even with the same language and cultural background, the fMRI recordings do not perfectly match the associative memory content. This discrepancy will inevitably introduce noise. We hope dataset recording associative memory of subjects is made to better investigate associative memory in the human brain using language models.



\section*{Ethics Statement}
In this paper, we explore associative memory in the human brain listening to speech through large language models. 
The proposed \textit{Association} dataset is for non-profit research usage.
Experiments are conducted on public accessible cognitive dataset ``Narratives'' with the authorization from its respective maintainers. The dataset has been de-identified by providers and is used for researches only.
\section*{Acknowledgements}
This research is supported by the National Natural Science Foundation of China (No.62476127, No.62106105),  the Natural Science Foundation of Jiangsu Province (No.BK20242039), the Basic Research Program of the Bureau of Science and Technology (ILF24001), the Fundamental Research Funds for the Central Universities (No.NJ2023032), the Scientific Research Starting Foundation of Nanjing University of Aeronautics and Astronautics (No.YQR21022), and the High Performance Computing Platform of Nanjing University of Aeronautics and Astronautics.


\bibliography{anthology,custom}
\bibliographystyle{acl_natbib}

\appendix

\section{Implemention Details}
\subsection{Cortical Parcellation}
\label{cort}
The latest version of Destrieux atlas \citep{DESTRIEUX20101} is applied for cortical parcellation, which leads to 74 regions per hemisphere. To reveal the associative memory in the human brain listening to speech, nine related regions are selected for experiments, including inferior frontal sulcus (S\_front\_inf), superior frontal sulcus (S\_front\_sup), middle frontal gyrus (G\_front\_middle), superior frontal gyrus (G\_front\_sup), angular gyrus (G\_pariet\_inf-Angular), superior parietal lobule (G\_parietal\_sup),  inferior temporal gyrus (G\_temporal\_inf), inferior temporal sulcus (S\_temporal\_inf), middle temporal gyrus (G\_temporal\_middle).

\subsection{Models and Hyper-parameters}
For fMRI data, we apply the AFNI-nosmooth preprocessing step for the Narratives dataset. Analyses are conducted on cortical voxels projected onto the surface and morphed onto an ``fsaverage6'' template brain.
We use `RidgeClassifierCV' regressor from scikit-learn \citep{pedregosa2011scikit} to predict the continuous features and align language models to brain, with 10 possible penalization values log-spaced between $10^{-1}$ and $10^8$. The linear model is evaluated on held out data, using 20 cross-validation for averaged score across all subjects and 5 cross-validation for brain score of each subject. 
For language models, we choose the small version of GPT-2 and LLaMA-2 with 7B parameters from Huggingface\footnote{\url{https://huggingface.co}}. The supervised finetuning of LLaMA-2 with LoRA or frozen layers is trained for 2 epochs with $10^{-4}$ learning rate. All experiments are conducted on NVIDIA A100-80G GPUs.

\subsection{Associative Memory Data Augmentation and The \textit{Association} Dataset}
\label{dataset}
In the data augmentation process of simulated associative memory, ten annotators are hired to make both word-level and sentence-level annotation. Annotators are asked to write down what they associate when receiving certain text stimuli that trigger associative memory. The hired annotators are Asian undergraduate students with English as their second language. Seven of them are male and three are female. Each annotator is assigned with two or three articles for labeling and is paid about 40 dollars. The annotators are informed that the data will be used for non-profit research. 

We also make GPT-4 version of data augmentation with the assistance of \texttt{gpt-4-1106-preview} API. The instruction tuning dataset \textit{Association} contains 1000 training samples with encouraging associative memory prompts and word-level association responses. It's composed of sentences from filtered stories of ``Narratives'' and sentences randomly picked from ROCStories dataset \citep{mostafazadeh2016corpus}. The \textit{Association} dataset is annotated with the help of GPT-4 through \texttt{gpt-4-1106-preview} API, more examples are shown in Appendix \ref{sec:appendix}.

\section{Case Study}
\label{sec:appendix}
In this part, we will take a deeper look into how data augmentation with associative memory is performed, and how the instruction tuning dataset \textit{Association} is made. More examples and cases are given and analyzed.

\subsection{Data Augmentation}
Table \ref{tab2} shows four examples of data augmentation. Four different augmentation methods are applied, including word-level augmentation by GPT-4 and human annotators, sentence-level augmentation by GPT-4 and human annotators. Since GPT-4 generates association content every three or four sentences, while human annotators add association stuff according to their ideas, the places of data augmentation are different in most cases. To facilitate a more effective comparison, we present the sentences that have been flagged by both GPT-4 and human annotators below.

\begin{table*}
  \centering
    \begin{tabular}{lcl}
    \toprule
    \multicolumn{1}{c}{\textbf{Original Sentences}}
          & \multicolumn{1}{c}{\textbf{Method}} & \multicolumn{1}{c}{\textbf{Data Augmentation}} \\
    \midrule
    \multirow{4}{*}{\makecell[l]{This is Los Angeles. And it's the\\ height of summer. In a small \\bungalow off of La Cienega, Clara \\serves homemade chili and chips \\in red plastic bowls -- wine in blue \\plastic.}}
        &word-level, GPT-4       & \makecell[l]{heat, bustling, cozy, spicy, casual, \\colorful} \\ \cmidrule{2-3}
          & word-level, human      & hot, comfortable\\ \cmidrule{2-3}
          & sentence-level, GPT-4      & \makecell[l]{The sun blazes down on a cozy home\\ in LA where a casual summer \\gathering unfolds.} \\ \cmidrule{2-3}
          & sentence-level, human      &  \makecell[l]{Clara uses plastic bowls of different\\ colors to make thing in a bungalow.}\\
    \midrule
    \multirow{4}{*}{\makecell[l]{Louis when I first started here. \\People told him, "Oh no, no she\\is white man, she's white, she \\sounds white she's white," and he, \\convinced, having never met me, \\that I was black. Well as it turns \\out, he was right.}}& word-level, GPT-4       &\makecell[l]{debate, racial identity, assumptions,\\ voice, community perceptions, \\prejudice, correctness, self-awareness,\\ revelation} \\ \cmidrule{2-3}
          & word-level, human      & \makecell[l]{debate, truth, revelation, surprise,\\ race}\\ \cmidrule{2-3}
          & sentence-level, GPT-4      & \makecell[l]{In a St. Louis debate about my ethnicity,\\ a stranger's conviction about my race\\ challenged the assumptions tied to my\\ voice,and he was correct. }\\ \cmidrule{2-3}
          & sentence-level, human      & \makecell[l]{Debates about the author being black\\ and white were going on long before\\ he came to St Louis Missouri\\ community.}\\
    \midrule
    \multirow{4}{*}{\makecell[l]{Jane named her Lucy and brought\\ her home on a commercial airline,\\ carried in a bassinet, her face \\covered with a lacy blanket. We \\were blissfully unaware of the \\complexities we were creatin on\\ the day Lucy came home. So \\the baby was a day or two old.}}& word-level, GPT-4       & \makecell[l]{Lucy, adoption, chimpanzee, travel,\\ naivety, complexities, infancy,\\ integration, new beginnings} \\ \cmidrule{2-3}
          & word-level, human       & \makecell[l]{Lucy, home, expectation, experiments} \\ \cmidrule{2-3}
          & sentence-level, GPT-4      & \makecell[l]{Lucy's journey veiled in the innocence of\\ infancy and a lacy blanket, commenced\\ with a flight to an uncharted life, while \\the Temerlins remained oblivious to the \\intricate future unfolding from their \\decision.} \\ \cmidrule{2-3}
          & sentence-level, human      &  \makecell[l]{Lucy would increase the complexity\\ of the experiments but I think she will\\ make it.}\\ 
    \midrule
    \multirow{4}{*}{\makecell[l]{Um, there's a lot of guys in army \\gear, um, shooting. It's very \\chaotic, can't really make out any \\faces, or, really people. Um, and\\ then, so, that, uh, cuts out really \\quickly and you see this man \\kind of like start out of the bed.}}&word-level, GPT-4       & \makecell[l]{battle, soldiers, military, chaos, abrupt\\ anonymous, awakening} \\ \cmidrule{2-3}
          & word-level, human      & \makecell[l]{soldier, war, army, grass field, chaotic}\\ \cmidrule{2-3}
          & sentence-level, GPT-4      & \makecell[l]{The man wakes up abruptly, haunted\\ by the chaos of battle.} \\ \cmidrule{2-3}
          & sentence-level, human      &  \makecell[l]{This seems to be the case with the\\ wars in the middle east.} \\
    \bottomrule
    \end{tabular}
  \caption{Examples of word-level and sentence-level data augmentation with associative memory.} 
  \label{tab2}
\end{table*}

\subsection{Instruction Tuning Dataset}

\begin{table*}
\centering
\begin{tabular}{llc}
\toprule
\multicolumn{2}{c}{\textbf{Input}} & \multirow{2}{*}{\textbf{Output}} \\
     \cmidrule(r){1-2}    \multicolumn{1}{c}{\textbf{Instruction}} & \multicolumn{1}{c}{\textbf{Story Paragraph}}\\ 
\midrule
     \makecell[l]{I'll give you some sentences, you \\ have to perform related association\\ with words.}        &      \makecell[l]{Sheldon slowly walked into the \\restaurant, eying the decor suspiciously.\\ His roommate Leonard pushed past him\\ and asked the hostess for a table for\\ two. As they were led to their chairs,\\ Sheldon began to protest yet again.}             &     \makecell[l]{quirky, cautious,\\ skeptical,\\ friends, dining,\\ impatient}                      \\ \midrule
        \makecell[l]{Given some sentences, you are\\ supposed to make related\\ associations and output words.}    &    \makecell[l]{You know, I think I may have\\ misjudged this restaurant. I won’t\\ go out on a limb, but I think we may\\ be looking at my new Tuesday\\ hamburger.}                &      \makecell[l]{surprise,\\ reconsideration,\\ hamburger,\\ potential favorite}                      \\ \midrule
        \makecell[l]{Given a batch of sentences,\\ you need to execute the process\\ of interlinking them based on\\ their relevance with words.}    &    \makecell[l]{He zipped up Barney’s bag and handed\\ it back to him. Quinn followed Barney\\ down the concourse in total confusion.\\ Magic trick? Why wouldn’t he tell\\ her what was in the box? She tried to\\ interrogate him as they sat in front of \\the gate, but he refused to spill the beans.}                &    \makecell[l]{mystery, secrecy,\\ curiosity, travel,\\ frustration,\\ companionship}                        \\ \midrule
        \makecell[l]{You need to engage in divergent\\ thinking based on the sentences\\ I provide, and give me whatever\\ words comes to your mind.}    &      \makecell[l]{That's fair, that's what we charge in\\ our country. After waiting for their\\ turn to board, they marched down the\\ jetway and onto the plane. George\\ struggled to get into his window seat \\and fit his bag down by his feet.}              &              \makecell[l]{equality, travel,\\ patience, boarding,\\ cramped, luggage,\\ discomfort}              \\ \midrule
        \makecell[l]{You will get a set of sentences,\\ and you need to associate some\\ related content with words.}    &   \makecell[l]{Vinny poked at it with his fork. What's\\ this over here? The cook looked at him\\ in disbelief. You've never heard of grits?\\ Sure, sure, I've heard of grits, I've just\\ never actually… seen a grit before.\\ Go ahead honey, aren't you going to \\try it? You first, she said with a smile.}                 &     \makecell[l]{curiosity,\\ skepticism,\\ southern cuisine,\\ breakfast, humor}                       \\ \midrule
        \makecell[l]{Given some sentences, you are\\ supposed to make related\\ associations and output words.}    &      \makecell[l]{Anna was filling her bird feeders. But\\ a chunk of suet fell onto the ground.\\ Her dog rushed over and lapped it up!\\ Anna was astonished. She had no idea\\ dogs loved bird food!}              &       \makecell[l]{surprise, dogs,\\ birds, feeding,\\ accidental,\\ curiosity}                     \\ \midrule
        \makecell[l]{Human tend to think relative\\ stuff when receiving text\\ information. Imagine you're\\ human and expand the following\\ sentences with words.}    &    \makecell[l]{Sam's dog Rex escaped from their yard.\\ Sam was distraught. He went out calling\\ for Rex. Then he saw Rex come running\\ up the street! Sam was so relieved,\\ he almost cried!}                &       \makecell[l]{worry, search,\\ reunion, joy,\\ pet, relief}                     \\ 
\bottomrule
\end{tabular}
\caption{Training samples randomly picked from the \textit{Association} dataset.}
\label{tab:tab1}
\end{table*}

The \textit{Association} dataset consists of input and output pairs as training samples. As shown in Table \ref{tab:tab1}, input content is made up of instruction and story paragraph. The instruction is prompts encouraging models to generate associative content, and the story paragraph contains sentences extracted from stories in ``Narratives'' dataset and ROCStories dataset. The output is word-level associated content. Table \ref{tab:tab1} displays some samples randomly selected from \textit{Association} dataset.

\end{document}